\documentclass[letterpaper]{article}
\usepackage{aaai}
\usepackage{times}
\usepackage{helvet}
\usepackage{courier}
\usepackage{CJK}
\usepackage{multirow}
\usepackage{amsmath}
\usepackage{algorithm}  
\usepackage{algorithmic}
\usepackage{graphicx}
\usepackage{bm}
\usepackage{booktabs}

\begin{document}
\title{
Dial2Desc: End-to-end Dialogue Description Generation
}
\author{Haojie Pan\textsuperscript{1},
Junpei Zhou\textsuperscript{2},
Zhou Zhao\textsuperscript{2},
Yan Liu\textsuperscript{3}, 
Deng Cai\textsuperscript{1},
Min Yang\textsuperscript{4}\\
\textsuperscript{1}State Key Lab of CAD\&CG, Zhejiang University\\ 
\textsuperscript{2}College of Computer Science, Zhejiang University \\
\textsuperscript{3}University of Southern California, \textsuperscript{4} Chinese Academy of Sciences \\
\{hjpan, jpzhou, zhaozhou\}@zju.edu.com, yanliu.cs@usc.edu, dengcai@gmail.com, min.yang@siat.ac.cn}

\maketitle
\begin{abstract}
We first propose a new task named Dialogue Description(\textit{Dial2Desc}). Unlike other existing dialogue summarization tasks such as meeting summarization, we do not maintain the natural flow of a conversation but describe an object or an action of what people are talking about. The \textit{Dial2Desc} system takes a dialogue text as input, then outputs a concise description of the object or the action involved in this conversation. After reading this short description, one can quickly extract the main topic of a conversation and build a clear picture in his mind, without reading or listening to the whole conversation. Based on the existing dialogue dataset, we build a new dataset, which has more than one hundred thousand dialogue-description pairs. As a step forward, we demonstrate that one can get more accurate and descriptive results using a new neural attentive model that exploits the interaction between utterances from different speakers, compared with other baselines.
\end{abstract}

\section{introduction}
Recently, a lot of novel techniques have been proposed to help people consume a large amount of text/audio data from the Internet or Daily life. Researchers and companies have successfully applied opinion mining or summarization into product reviews, news articles, scientific articles, etc. However, very little attention has been given so far to help people consume dialogue records which are generated every day. It is clear that automatic summarization of dialogues can be of benefit in dealing with this overwhelming amount of interactional information\cite{MurrayC08}. 

\begin{table}[!t]
    \caption{An example of Dial2Desc}
    \vspace{0.1cm}
    \centering
    \begin{tabular}{|l|}
    \hline
    \textbf{Dialogue}: \\
    A: Is this in color. \\
	B: No , it 's black and white. \\
	A: Does it look like an old picture. \\
	B: Yes , i think so. \\
	A: How old do you think the man is. \\
	B: It looks like a young boy and he is what makes me think \\the picture is older , but the picture is not really old. \\
	A: Do you see more than 1 cow. \\
	B: 2 cows. \\
	A: Is the boy wearing overalls. \\
	B: No , he 's wearing a plaid short sleeved shirt and \\long pants and regular shoes. \\
	A: How about a hat. \\
	B: No. \\
	A: Has he already started to milk. \\
	B: He is attaching the milking apparatus to the cow. \\
	A: Is he sitting down. \\
	B: He is squatting. \\
	A: What color is the cow. \\
	B: Looks like it would be brown and white. \\
	A: Are they inside a barn. \\
	B: Inside a milking facility. \\
	\hline
	\textbf{Description}: a man prepares to milk a dairy cow.\\
    \hline
    \end{tabular}
\end{table}

Previous conversation summarization works including using extractive approaches \cite{xie2008evaluating,riedhammer2010long} or abstractive approaches \cite{oya2014template,banerjee2015multi,ShangGuokan18} on meeting summarization, which generates summaries that allow people to prepare for an upcoming meeting or review the decisions of a previous group. There is also an increasing research interest in other conversation summarization, such as conversational telephone speech, broadcast news, lectures, and e-mails. Those tasks are mostly focusing on stating the events in a natural flow of raw conversations and can hardly string events together to form the thematic abstracts of the whole transcripts. However, most of the human conversations are involving some specific objects or actions, and speakers may have a clear picture of what they are talking about, which can not be described without stringing speakers' statements together.

In this work, we proposed a novel task named \textit{Dial2Desc}, which is a variant of conversation summarization. Unlike the previous works mentioned above, we pay more attention to a higher-abstractive-level of the dialogues, instead of maintaining the natural flow of the given conversations. The target of our task is to describe an object or an action of what speakers are talking about in a dialogue transcript. 

In the last several years, we can see deep learning has boosted the development of summarization in written text, such as news articles. Researchers apply modern neural networks with attention mechanism on abstractive summarization \cite{RushCW15,ChopraAR16,NallapatiZSGX16,SeeLM17,paulus2018a} and reach the state-of-the-art results on several datasets. The availability of large-scale parallel summarization dataset and powerfulness on the representation of deep learning push this task into a new stage, where one can achieve good results without doing any complex preprocessing procedures. However, in conversation summarization, researchers tend to build unsupervised models and involve manual rules because of the lack of such high-quality datasets.

Inspired by current \textit{Image2Text} works in computer vision, we find there are high-valued image-text-mixed datasets in Image Caption tasks, which is to use salient visual information into descriptive languages\cite{XuBKCCSZB15}, and VisualDialog, which requires an AI agent to hold a meaningful dialog with humans in natural, conversational language about visual content\cite{Abhishekvisdial17}. We now can use an image as a bridge to align the connected dialogue and description(that is where the name \textit{Dial2Desc} comes from). One can intuitively find that this image is exactly what speakers are talking about or a picture in mind during the conversation, and the captions are the higher-abstractive-level way to describe the dialogues.

We collect 122,621 dialogues from VisualDialog Dataset\cite{Abhishekvisdial17} and corresponding captions from the COCO dataset \cite{LinMBHPRDZ14} to build our final dataset, which enables us to develop more advanced neural models for this task. One of the examples is shown in Table 1.

Directly applying neural models proposed for written text summarization is not a good idea, since spoken conversation languages have some additional issues, e.g., maintaining cross-speaker coherence\cite{Zechner01}. Most of the neural abstractive summarization models are based on sequence-to-sequence framework\cite{seq2seq2014}, which consider the whole input article as a source sequence and use recurrent ways or hierarchical ways to encode it. However, the interactions between speakers and the flows of dialogue play a more important role in dialogue modeling, which is overlooked. 

To address this problem, we propose a novel neural encoder, which use co-attention mechanism and dense connection to enable interactions between speakers and use the transformer framework to maintain the message passing during dialogue turns. And we also apply the transformer as our decoder. The experiment results show that our encoder plus transformer decoder can achieve the highest performance over other summarization baselines.

In summary, we make the following contributions:
\begin{enumerate}
    \item A novel task named \textit{Dial2Desc}, which address the generation of higher-abstractive-level description over the objects or the actions that people are talking about.
    \item A large-scale dataset built from existing public dialogue datasets for our task. 
    \item A novel neural attentive model that exploits the interaction between utterances from different speakers.
\end{enumerate}

\section{Related Work}
\subsection{Image caption and visual Dialogue}
Image caption has been widely studied\cite{VinyalsTBE15,YouJWFL16,AndersonFJG17}. In general, researchers use Convolutional Neural Network to encode a given picture and then use Recurrent Neural Network, especially its variant Long-Short Term Memory \cite{HochSchm97} to decode this semantic representation. Visual Dialog \cite{Abhishekvisdial17}, is a task that when given an image, a dialog history, and a question about the image, the agent has to ground the question in the image, infer context from history, and answer the question accurately \cite{Abhishekvisdial17}. In this work, we build a bridge between this two tasks to build our dataset, since one of image caption datasets and the VisualDialog dataset\footnote{https://visualdialog.org} are both evolved from MSCOCO dataset\footnote{http://cocodataset.org/\#home}.

% ,KarpathyL15,DevlinCFGDHZM15,WuSLDH16
\begin{figure*}[!t]
\centering
\includegraphics[width=16cm]{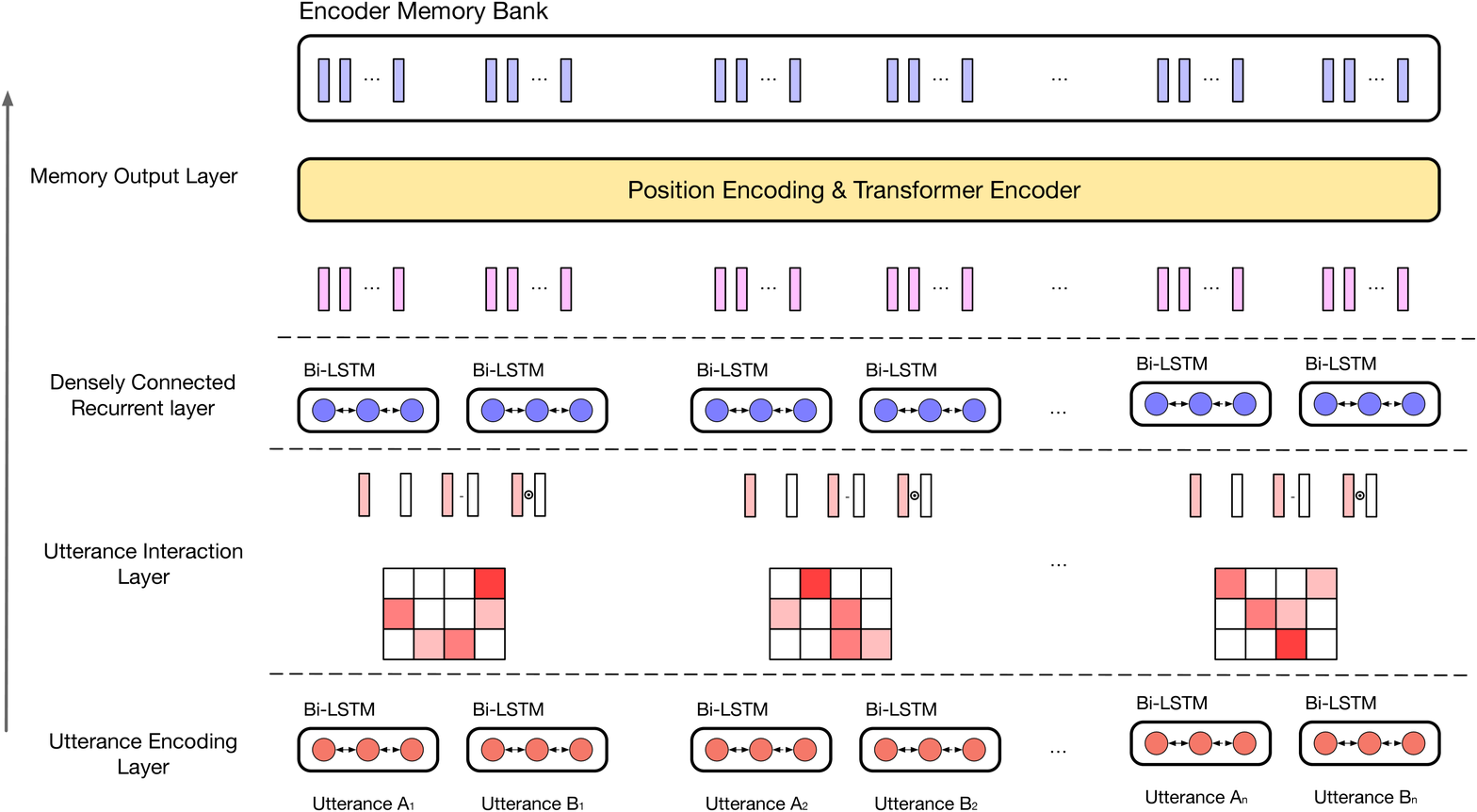}
\caption{Enhanced Interaction Dialogue Encoder}
\end{figure*}

\subsection{Conversation summarization}
Recently, there is an increasing research interest in speech summarization. \cite{Zechner01} explored aspects of speech transcripts, e.g. disfluencies, to generate summaries for conversational telephone speech. \cite{MaskeyH05} explored supervised and unsupervised approaches with different kinds of features on broadcast news. \cite{ZhangCF07} used extractive methods on lecture speech transcripts. On mail thread summarization, \cite{NenkovaB03} proposed a method to generate a summary for the first two levels of the thread discussion. \cite{RambowSCL04} used a machine learning technique and included features related to the thread as well as features of the email structure such as the position of the sentence in the thread, number of recipients, etc. And on meeting summarization, \cite{xie2008evaluating} treated this task as a binary classification problem.\cite{riedhammer2010long} analyzed and compared two different methods for unsupervised extractive meeting summarization. Oya \cite{oya2014template} leveraged the relationship between human-authored summaries and their source meeting transcriptions to select the templates for generating abstractive summaries for meetings. Banerjee \cite{banerjee2015multi} generated abstractive summaries by fusing important content from several utterances with the dependency graph. \cite{ShangGuokan18} combined the strengths of multiple recent approaches introduce a novel graph-based framework for unsupervised abstractive meeting summarization. Our task, different with these tasks, address higher-abstractive-level of dialogues, which targets on describing what people are talking about instead of stating events of conversations.

\subsection{Neural attentive models and summarization}
Neural attentive models play important roles in many tasks. such as  machine translation\cite{seq2seq2014}, text matching\cite{ChenZLWJI17}, and question answering\cite{HermannKGEKSB15}. Attention mechanisms \cite{BahdanauCB14} make these models more performant and scalable, allowing them to look back at parts of the encoded input sequence while the output is generated. 

Researchers introduce those models into text summarization \cite{RushCW15,ChopraAR16,NallapatiZSGX16}. \cite{NallapatiZSGX16} used different attention and pointer functions on the CNN and Daily Mail datasets combined. \cite{SeeLM17} developed an abstractive summarization model on this dataset with an extra loss term to increase temporal coverage of the encoder attention function. And \cite{paulus2018a} used intra-attention and reinforcement learning to boost the results of summarization. In this work, we adapt neural attentive models into conversation summarization scenario and exploit the interaction between utterances from different speakers.

\section{Our Approach}
In this section, we describe the architecture of our neural model, including an enhanced interaction dialogue encoder and transformer-pointer generator.
\subsection{Position Encoding and Multi-head Attention}
First, we introduce some background on position encoding and multi-head attention \cite{VaswaniSPUJGKP17}, which are the building block for our model.

Position encoding is designed for models contains no recurrence or convolution, to make use of the order of a given sequence. We use sine and cosine functions of different frequencies to build our Position Encoder(PE):

\begin{eqnarray}
	PE_{(pos,2i)} = sin(pos/10000^{2i/d_{model}}) \\
	PE_{(pos,2i+1)} =cos(pos/10000^{2i/d_{model}})
\end{eqnarray}
where $pos$ is the position and $i$ is the dimension. That is, each dimension of the positional encoding corresponds to a sinusoid. The wavelengths form a geometric progression from $2\pi$ to $10000 \cdot 2\pi$. This function would allow the model to easily learn to attend by relative positions \cite{VaswaniSPUJGKP17}.

To introduce Multi-head attention, we start from the scaled dot-product attention. Given a query $q_i \in R^d$ from all $T$ queries, a set of keys $k_t \in R^d$ and values $v_t \in R^d$ where $t = 1, 2, ..., T$ , the scaled dot-product attention outputs a weighted sum of values $v_t$, where the weights are determined by the dot-products of query $q$ and keys $k_t$. In practice, we pack $k_t$ and $v_t$ into matricies $K = \{ k_1, k_2, ..., k_T \} $ and $V = \{ v_1, v_2, ..., v_T \}$, respectively. The attention output on query q is:
\begin{equation}
	A(q_i, K, V) = V\frac{K^Tq_i/\sqrt{d}}{\sum_{t=1}^Texp(k_t^Tq_i/\sqrt{d})}
\end{equation}
The multi-head attention consists of $H$ paralleled scaled dot-product attention layers called "head", where each "head" is an independent dot-product attention. The attention output from multi-head attention is as below:
\begin{eqnarray}
	MA(q_i, K, V) = [head_1, head_2,...head_H]W^O \\
	head_j = A(W_j^Qq_i, W_j^KK, W_j^VV) 
\end{eqnarray}
where the projections are parameter matrices $W_j^Q \in R^{d_{model} \times d}$, $W_j^K \in R^{d_model \times d} , W_j^V \in R^{d_{model} \times d}$ and $W^O \in R^{Hd \times d_{model}}$.
Both formulations of $A$ and $MA$ is quite general, and it represents the common cross-module attention. If queries, keys, and values are all the same, it is called Self-attention \cite{VaswaniSPUJGKP17}.

\subsection{Enhanced Interaction Dialogue Encoder}
In this section, we describe our encoder which is composed of the following four components: (1) utterance encoding layer, (2) utterance interaction layer, (3) densely connected recurrent layer, (4) memory output layer, to encode the dialogue into a memory. We denote Dialogues as A/B speaker utterance pairs $\{(A_1, B_1), (A_2, B_2),..., (A_n, B_n)\}$, where $n$ is the number of turns in a dialogue. For each pair, $A = \{a_1, a_2, ..., a_{\ell_{A}} \}$ and $B = \{b_1, b_2,..., b_{\ell_{B}} \}$, where $\ell_{A}$ and $\ell_{B}$ are number of tokens of $A$ and $B$ (we omit the subscript of $A, B$ for convenience).
\subsubsection{utterance encoding layer} encodes word list into fixed context vector list. Firstly, we represent each pair as list of $d$ dimension word embeddings:
\begin{eqnarray}
	v_A = \{ v_{a_1}, v_{a_2},...,v_{a_{\ell_{A}}} \} ;~
	v_B = \{v_{b,1}, v_{b,2},...,v_{b_{\ell_{B}}} \} 
\end{eqnarray}
and we then employ bidirectional LSTM \cite{GravesFS05} for preserving sequential information of $A$ and $B$ (we skip the description of the basic chain LSTM due to the space limit):
\begin{eqnarray}
	h_{a_t} = biLSTM(v_{a_t}, h_{a_{t-1}}) \\
	h_{b_t} = biLSTM(v_{b_t}, h_{b_{t-1}}) 
\end{eqnarray}
and obtain utterance representations:
\begin{eqnarray}
\bm{h_a} = \{h_{a_1}, h_{a_2},...,h_{a_{\ell_{A}}} \};~
\bm{h_b} = \{h_{b_1}, h_{b_2},...,h_{b_{\ell_{B}}} \}
\end{eqnarray}

\subsubsection{utterance interaction layer} uses attention mechanism in Equation (3) to re-encode the contexts from interaction perspective. The local interaction information $s_{a_i}$ of the $i^{th}$ word $a_i \in A$, and $s_{b_i}$ of the $i^{th}$ word $b_i \in B$ is computed by using scaled dot-product attention as follows:

%\begin{eqnarray}
%	s_{a_i} = \sum_{j=1}^{\ell_B}\frac{exp(e_{i, j})}{\sum_{k=1}^{\ell_B}exp(e_{i,k})}h_{b_j} \\
%	s_{b_i} = \sum_{i=1}^{\ell_A}\frac{exp(e_{i, j})}{\sum_{k=1}^{\ell_A}exp(e_{k,j})}h_{a_j}
%\end{eqnarray}

\begin{eqnarray}
	s_{a_i} = A(h_{a_i}, \bm{h_b}, \bm{h_b});~
	s_{b_i} = A(h_{b_i}, \bm{h_a}, \bm{h_a})
\end{eqnarray}

And we further enhance the local interaction information by computing the difference and the element-wise product for the tuple $<\bm{h_{a}}, \bm{s_{a}}>$ as well as for $<\bm{h_{b}}, \bm{s_{b}}>$ \cite{ChenZLWJI17}. The difference and element-wise product are then concatenated with the original vectors, $\bm{h_{a}}$ and $\bm{s_{a}}$, or $\bm{h_{b}}$ and $\bm{s_{b}}$, respectively, then we get:

\begin{eqnarray}
	 \bm{\tilde{s}_a} = [\bm{h_{a}}; \bm{s_{a}}; \bm{h_{a}} - \bm{s_{a}}; \bm{h_{a}} \odot \bm{s_{a}} ] \\
	  \bm{\tilde{s}_b} = [\bm{h_{b}}; \bm{s_{b}}; \bm{h_{b}} - \bm{s_{b}}; \bm{h_{b}} \odot \bm{s_{b}}]
\end{eqnarray}

\subsubsection{densely connected recurrent layer} uses another biLSTM to enables us to build up higher-level representation. Instead of directly inputting the encoded hidden features from the last layer, we concatenate them with the word embeddings, to preserve the encoded hidden features until they reach to the uppermost layer and all the previous features work for prediction as collective knowledge \cite{HuangLMW17}:

\begin{eqnarray}
	h_{a_t}^{'} = biLSTM(x_{a_t}, h_{a_{t-1}}^{'}), x_{a_t}=[\tilde{s}_{a_t};v_{a_t}] \\
	h_{b_t}^{'} = biLSTM(x_{b_t}, h_{b_{t-1}}^{'}), x_{b_t}=[\tilde{s}_{b_t};v_{b_t}] 
\end{eqnarray}
%
%\subsubsection{context interaction layer} firstly encodes the series of encoded hidden features into fixed contexts $c_a$ for A and $c_b$ for B, then use the same way as Equation (9) or (10) to enhance the contexts. We employ structured self-attentive sentence embedding \cite{LinFSYXZB17} to encode $ \bm{h_{a}^{'}} $ and $ \bm{h_{b}^{'}} $, separately:
%\begin{eqnarray}
%	\alpha_{a, self} = softmax(w_{s_2}tanhW_{s_1}\bm{h_{a}^{'T}}) \\
%	\alpha_{b, self} = softmax(w_{s_2}tanhW_{s_1}\bm{h_{b}^{'T}})
%\end{eqnarray}
%
%Here $W_{s_1}$is a weight matrix with shape of $d_{self}$-by-$h$. And $w_{s_2}$is a vector of parameters with size $d_{self}$, where $h$ is the size of hidden state,  $d_{self}$ is a hyperparameter we can fine-tune afterwards. Then we use weighted summation to get fixed contexts:
%\begin{equation}
%	c_a = \sum_{i=1}^{\ell_a}\alpha_{a, self, i}h_{a_i}^{'}, c_b = \sum_{i=1}^{\ell_b}\alpha_{b, self, i}h_{b_i}^{'}
%\end{equation}
%and we enhance the contexts to get final representation of A/B pairs:
%\begin{equation}
%	\tilde{c} = [c_a; c_b; c_a - c_b; c_a \odot c_b]
%\end{equation}
%
%\subsubsection{turns aggregation layer} uses another biLSTM to gain some dependency between each turns. we put $\title{c}_{t}$ as input at time step $t$:
%\begin{equation}
%	\tilde{h}_t = biLSTM(c_t, \tilde{h}_{t-1})
%\end{equation}

\begin{figure}[!t]
\centering
\includegraphics[width=8cm]{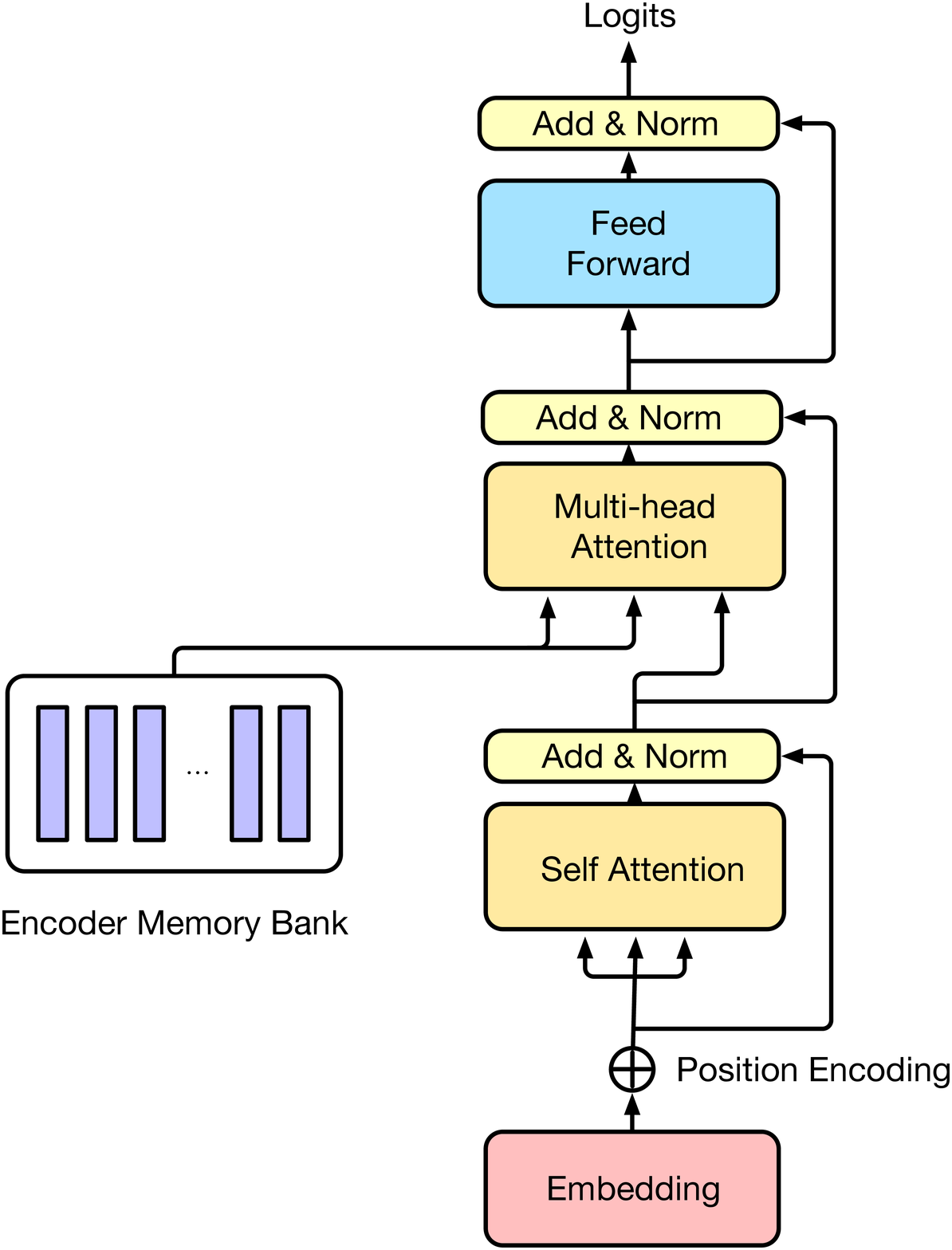}
\caption{Transformer Decoder}
\end{figure}

\subsubsection{memory output layer}
We concatenate all the $h_{a_t}^{'}$ and $h_{b_t}^{'}$ in each turn $k$ into a sequence of hidden memories $M$. Then we follow the work \cite{VaswaniSPUJGKP17} to apply position encoding into $M$ and apply one transformer layer to output a turns-aware encoder memory bank $M'$. The transformer layer contains two sublayers: (1) a self-attention layer, where we take the output of the previous layer as queries, keys, and values and employ multi-head attention mechanism.  (2) a simple, position-wise fully connected feed-forward network which is applied to each position separately and identically:
\begin{equation}
    FFN(x) = max(0, xW_1 + b_1 )W_2 + b_2
\end{equation}
In addition, we employ a residual connection \cite{HeZRS16} around each of the two sub-layers, followed by layer normalization \cite{BaKH16}. That is, the output of each sub-layer is $LayerNorm(x + Sublayer(x))$, where $Sublayer(x)$ is the function implemented by the sub-layer itself. The overall encoder is shown in Figure 1.

\subsection{Transformer-pointer Generator}
We also use the transformer as the basic block of our description decoder, as shown in Figure 2. Then we use pointer network \cite{VinyalsFJ15} to generate final outputs, as it allows both copying words via pointing and generating words from a fixed vocabulary.

We first put decoder inputs into an embedding layer, which is similar to encoder embedding layer. And we use position encoding in Equation (1) and (2) to make use of sequential information. For one decoder-transformer layer, there is an extra layer compared with encoder transformer layer: a context multi-head attention layer where the queries come from the previous decoder layer, and the memory keys and values come from the output of the encoder, $M'$. After multiple transformer layers, we get decoder state $s_t$ for each decoder timestep $t$. 
Then we put $s_t$ into one linear layer to produce the vocabulary distribution $P_v$

\begin{equation}
    P_v = softmax(V^Ts_t + b)
\end{equation}
where $V$ and $b$ are learnable parameters.

For each decode step $t$, we take the attention weights of the second sub-layer of the last transformer layer as the encoder memory attention distribution $a^t$. The generation probability $pgen \in [0, 1]$ for timestep $t$ is calculated from the decoder state $s_t$:

\begin{equation}
    pgen = \sigma(w_{ptr}^Ts_t + b_{ptr})
\end{equation}
where vectors $w_{ptr}$and scalar $b_{ptr}$ are learnable parameters and σ is the sigmoid function. Then we use $p_{gen}$ to choose between generating a word from the vocabulary by sampling from $P_v$, or copying a word from the input sequence by sampling from the attention distribution $a^t$. For each dialogue, we get an extended vocabulary from the union of the vocabulary and all words appearing in the source dialogue. We obtain the following probability distribution over the extended vocabulary:
\begin{equation}
	P(w) = p_{gen} P_v(w) + (1 - p_{gen} ) \sum_{i:w_i = w} a_i^t
\end{equation}
This pointer generator models have advantages of producing OOV words compared to other seq2seq models which are restricted to their pre-set vocabulary.

During training, the loss for timestep $t$ is the negative log likelihood of the target words $w_t^∗$ for that timestep:
\begin{equation}
	loss_t = -log(w_t^*)
\end{equation}
and the overall sequence loss is:
\begin{equation}
	loss = \frac{1}{T}\sum_{t=1}^Tloss_t
\end{equation}

\section{Experiments}
In this section, we describe the dataset, experimental setup, evaluation metrics and the results of our experiments.

\subsection{Dataset}
\textit{Dial2Desc} dataset is based on two public data resources:

\textbf{VisDial} \cite{Abhishekvisdial17}: Visual Dialog is a task that requires an AI agent to hold a meaningful dialog with humans in natural, conversational language about visual content. Using images from MSCOCO dataset\cite{LinMBHPRDZ14}. They paired 2 workers on AMT to chat with each other in real-time to build dialogues that have (1) temporal continuity, (2) grounding in the image, and (3) mimic natural ‘conversational’ exchanges. VisDial v0.9 has been released and contains 1 dialog with 10 question-answer pairs on ~120k images from COCO, with a total of ~1.2M dialog question-answer pairs. And every dialogue in VisDial has 10 turns.

\textbf{MSCOCO} \cite{LinMBHPRDZ14}: this dataset contains human annotated captions of over 120K images. Each image contains five captions from five different annotators. This dataset is a standard benchmark dataset for image caption generation task. In a majority of the cases, annotators describe the most prominent object/action in an image, which makes this dataset suitable for our setting.

We can find the dialogues from VisDial and captions from MSCOCO are from the same image set. The intuition is that when two speakers are talking about one object or scene, they may have a clear picture in their mind, which can be described using higher-level-abstractive captions. 

To build our \textit{Dial2Desc} dataset, we collected all the dialogues from VisDial, and find attached image captions from MSCOCO dataset. Then we selected distinct captions with their attached dialogues to create dialogue-to-description pairs. Finally, we got 122,621 training pairs. And we split them into train/dev/test set. Some statistics are shown in Table 2.

Furthermore, for test data, we collected 5 descriptions in total for each dialogue, to ensure the stability of evaluation.

\begin{table}[!t]
    \caption{An overview of \textit{Dial2Desc} dataset}
    \vspace{0.2cm}
    \centering
    \begin{tabular}{c|ccc}
    \multirow{2}*{~} & \multirow{2}*{\#samples} & mean  & mean\\
    ~ & ~ & \#dialog tokens & \#desc tokens \\
    \hline
    train & 98,256 & 122.8 & 10.59 \\
    dev & 12,282 & 122.7 & 10.6 \\
    test & 12,083 & 122.6 & 10.6 
    \end{tabular}
\end{table}

\subsection{Baselines}
We empirically find that some unsupervised summarization methods such as Maximal Marginal Relevance(MMR) cannot reach good results because our ground truth descriptions are so abstractive. So we compare our methods with several neural generative approaches as follows:

\begin{table*}[!t]
	\centering
    \caption{Performance comparison of the proposed method with other methods on \textit{Dial2Desc} dataset}
	\vspace{0.2cm}
    \begin{tabular}{lccccccc}
    \toprule
    ~ & BLEU-1 & BLEU-2 & BLEU-3 & BLEU-4 & ROUGE-L & METEOR & CIDEr \\
    \midrule
    Attn-Seq2seq(20k vocab) & 66.2 & 48.4 & 34.4 & 24.4 & 47.8 & 22.6 & 82.2 \\
    Attn-Seq2seq(28k vocab) & 65.0 & 47.7 & 33.8 & 23.9 & 46.7 & 22.3 & 80.5 \\
    PGN & 67.5 & 49.9 & 35.4 & 25.1 & 48.4 & 22.9 & 85.6 \\ 
    Onmt-brnn & 68.0 & 50.6 & 36.4 & 26.0 & 49.0 & 23.2 & 86.5 \\
    Onmt-transformer & 67.2 & 50.2 & 35.9 & 25.5 & 49.2 & 22.9 & 84.9 \\
    \textbf{Our model} & \textbf{69.6} & \textbf{53.1} & \textbf{39.0} & \textbf{28.4} & \textbf{50.6} & \textbf{24.2} & \textbf{94.0} \\
    \bottomrule
    \end{tabular}
\end{table*}

\begin{itemize}
    \item \textbf{Attn-Seq2seq} is a base model described in \cite{SeeLM17}, where the encoder is a single-layer bidirectional LSTM producing a sequence of encoder hidden states and the decoder is a single-layer unidirectional LSTM which exploit the information from the encoder hidden states via attention mechanism.
    \item \textbf{PGN} is a hybrid pointer-generator network that can copy words from the source text via pointing \cite{SeeLM17}. \footnote{Implementation of Attn-Seq2seq and PGN can be found in https://github.com/abisee/pointer-generator}
    \item \textbf{Onmt-brnn} is similar to PGN, which use a bidirectional LSTM to encode dialogues, a unidirectional LSTM and copy-mechanism to generate descriptions. However, the implementation details are slightly different from PGN.
    \item \textbf{Onmt-transformer} use transformer\cite{VaswaniSPUJGKP17} framework(which is the state-of-the-art model in machine translation) combined with a copy-generator. \footnote{Impletations of Onmt-brnn and Onmt-transformer can be found in https://github.com/OpenNMT/OpenNMT-py} 
%	\item \textbf{Our model} use a enhanced interaction dialogue encoder and transformer-pointer generator
\end{itemize}

\subsection{Experimental setup}
We use a vocabulary of 20k words, shared by both source(dialogues) and target(descriptions) for all of the models, to make use of copy-mechanism. For Attn-Seq2seq, we also try a larger vocabulary size of 28k (almost the same as the vocabulary of training data). And for all RNN-based models, 256-dimensional RNN hidden states and 128-dimensional word embeddings are applied. And we use Adagrad \cite{DuchiHS11} with learning rate 0.15 and an initial accumulator value of 0.1 to train these models. For Onmt-transformer and our model, both the dimension of RNN hidden states and word embeddings are set to 256. Following the work \cite{VaswaniSPUJGKP17}, we used the Adam optimizer \cite{KingmaB14} with $\beta_1 = 0.9, \beta_2=0.98$ and $\epsilon = 10^{-9}$, and the $warmup\_steps$ is set to 8000. Word embeddings are learned from scratch during training instead of using any pre-trained ones. 

During training and at test time we truncate the utterance of each turn to 20 tokens and limit the length of the summary to 5-15 tokens for both training and at test time. We use PyTorch to conduct all the experiments, and all the models are trained in GTX 1080Ti GPU with a batch size of 16 for RNN-based models and a batch size of 4096 for transformer-based models.
\subsection{Evaluation Metrics}
Same as image caption task, we use several unsupervised automated metrics for NLG(Natural Language Generation) to evaluate baselines and our model. Those metrics include BLEU  \cite{PapineniRWZ02}, ROUGE-L \cite{Lin2004ROUGEAP}, METEOR  \cite{LavieA07}, CIDEr \cite{VedantamZP15}.When computing CIDEr, we compute IDF values using the reference sentences provided to adapt our setting,  which is different with image caption task.
We use nlg-eval to conduct evaluation\footnote{https://github.com/Maluuba/nlg-eval}.

\begin{figure}[t]
\centering
\includegraphics[height=0.17\textwidth]{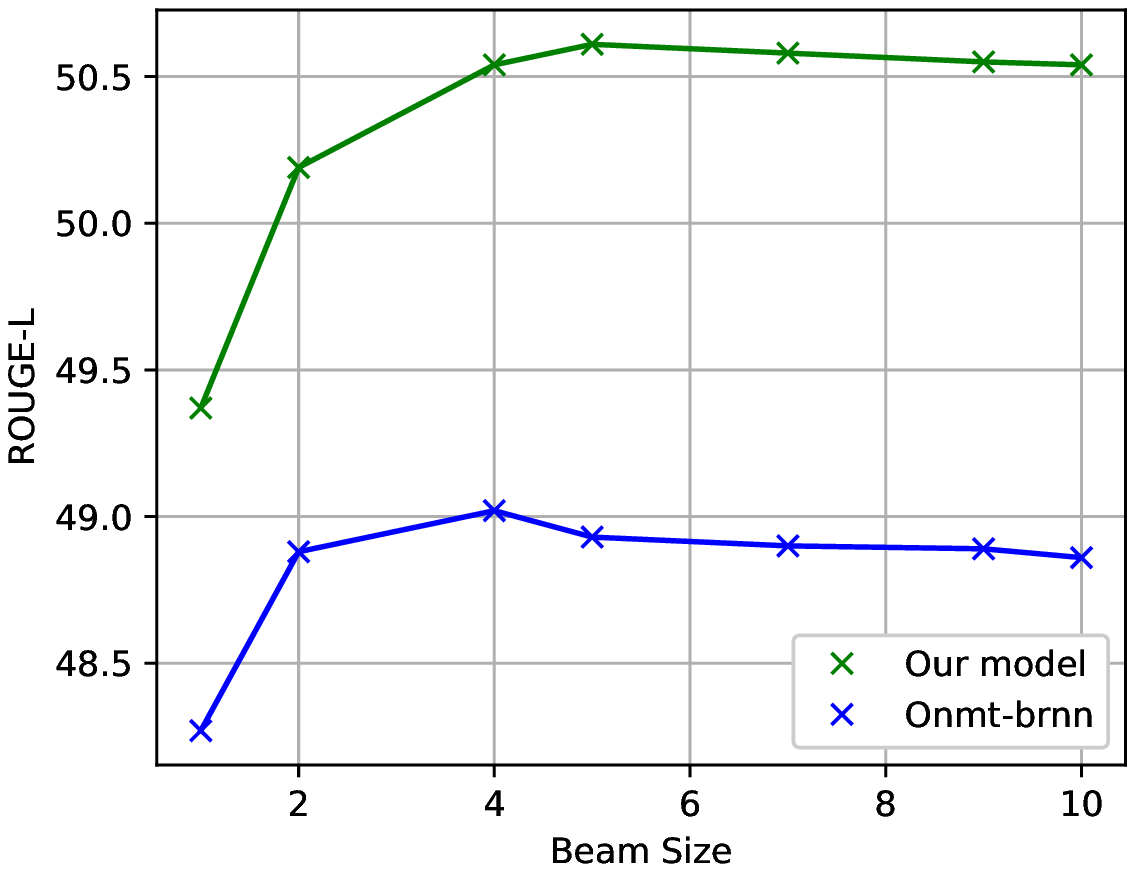}
\includegraphics[height=0.17\textwidth]{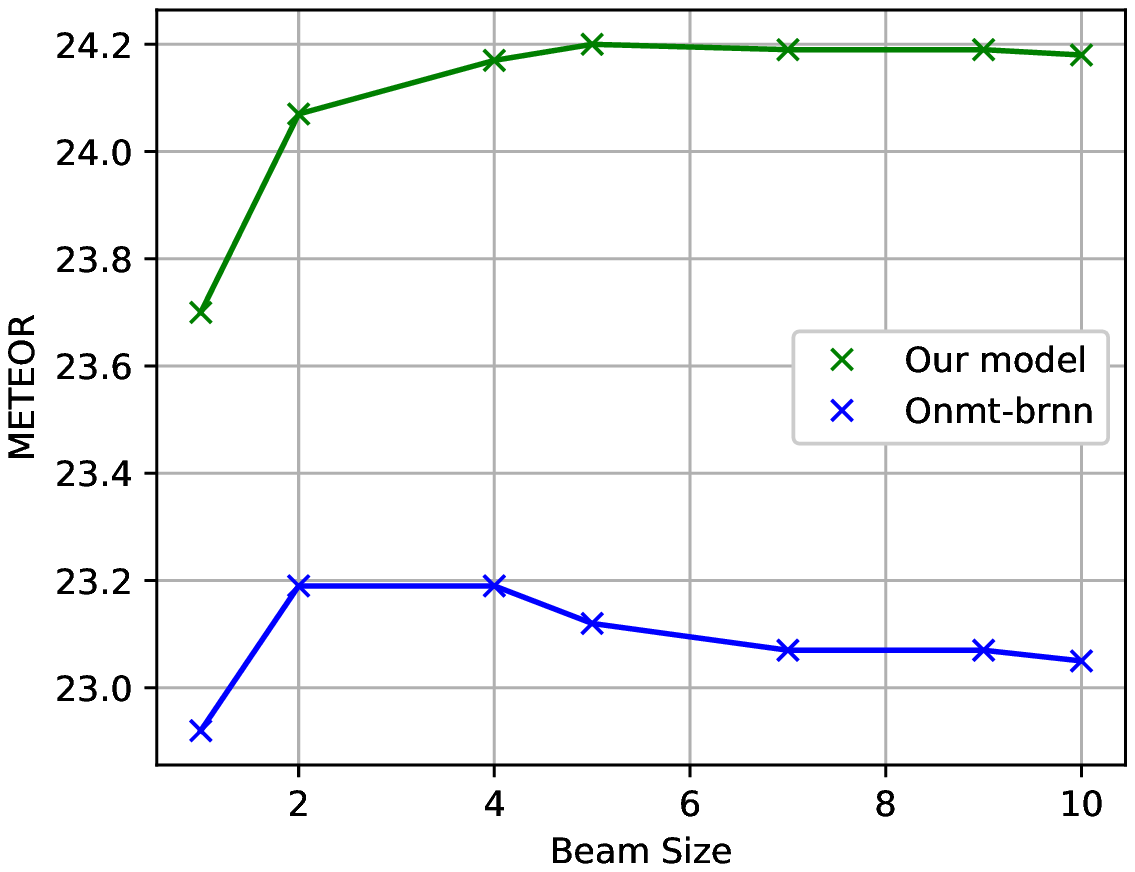}
\caption{The influence of beam search size K on the Onmt-brnn and our model}\label{fig:co-attn}
\end{figure}

\begin{figure}[t]
\centering
\includegraphics[height=0.11\textwidth]{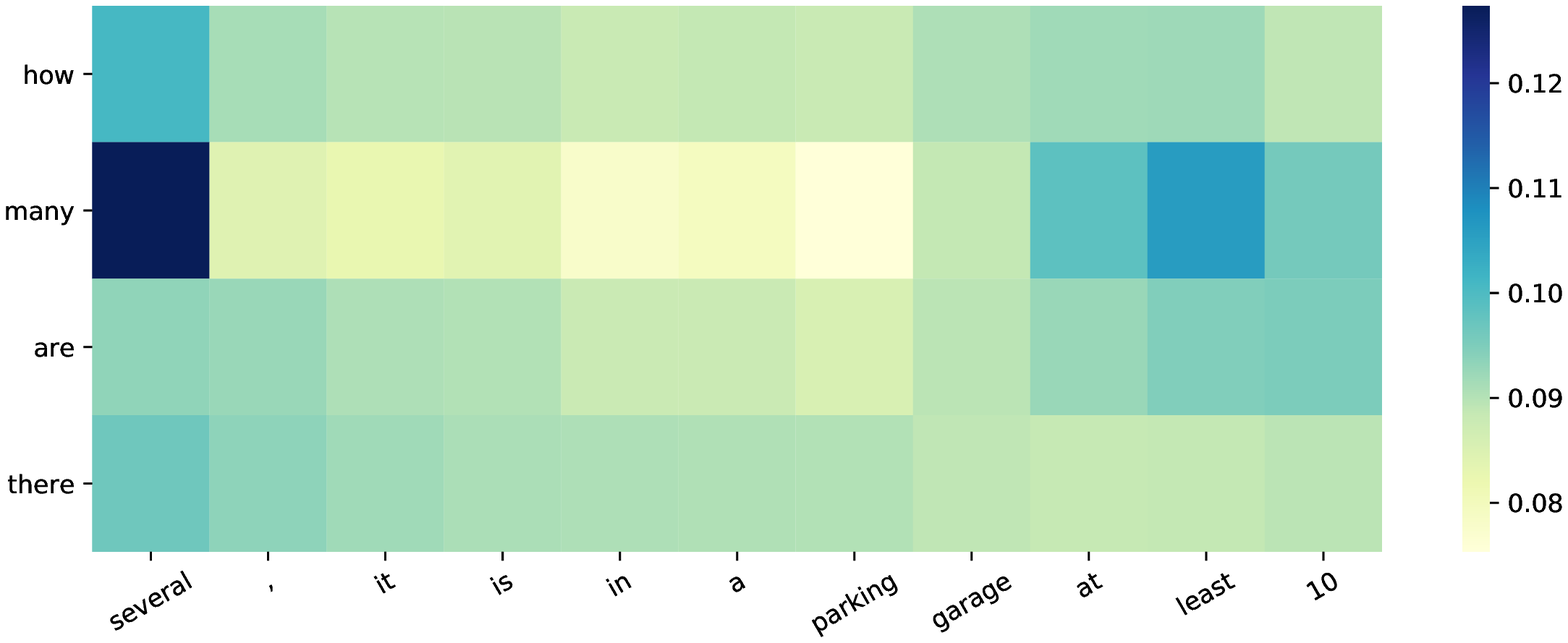}
\includegraphics[height=0.11\textwidth]{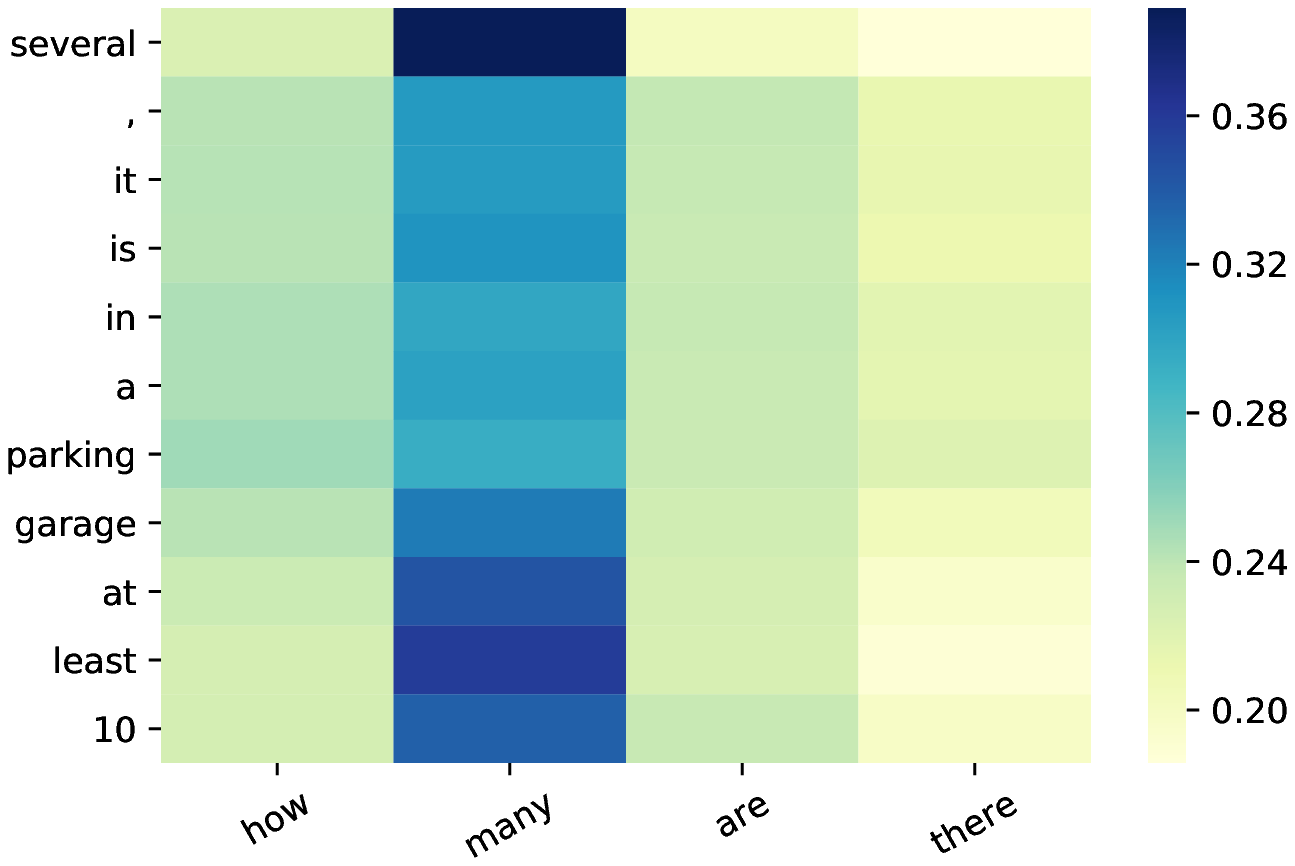}
\caption{Co-attention between utterences from two speakers. The summation of each row is equal to 1}\label{fig:co-attn}
\end{figure}

\begin{figure*}[t]
\centering
\includegraphics[height=0.48\textwidth]{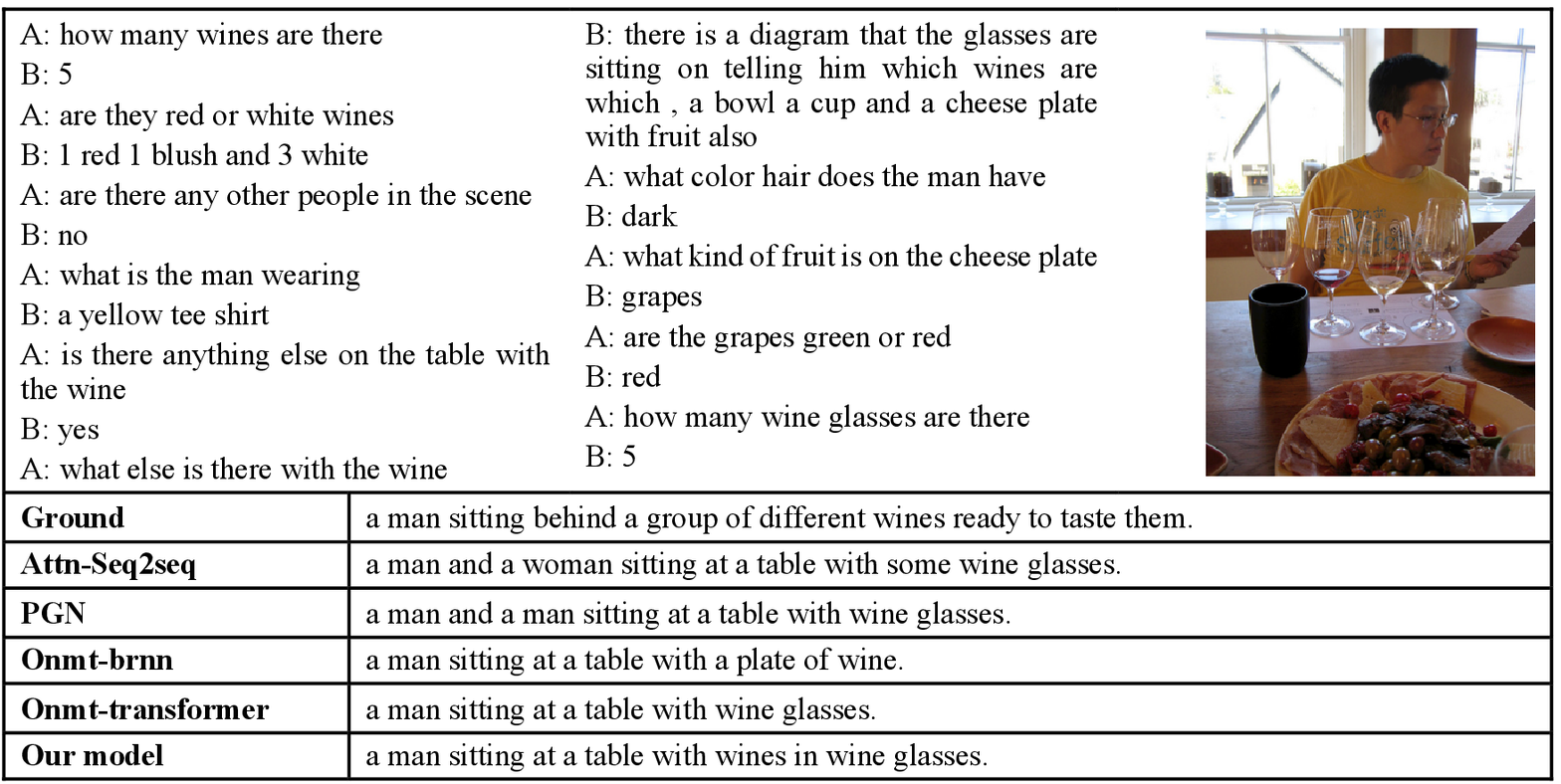}
\caption{A case study}\label{fig:case}
\end{figure*}

\subsection{Results}
We perform experiments on \textit{Dial2Desc} dataset, and report both qualitative and quantitative results of our approach.

\subsubsection{Quantitative Results}

The results of mentioned baselines and our model are listed in Table 3. As we can see, we have a significant improvement w.r.t.the baselines. We can find vocabularies is very important to those neural generative models. Attn-Seq2seq, which has no other way to generate OOV words, performs worst. And in fact, the larger vocabulary size does not seem to help. However, when copy mechanism is applied, models such as PGN, Onmt-brnn gain a very big improvement, compared with Attn-Seq2seq. Onmt-transformer with copy generator has also as good performance as RNN-based approaches. It reaches better scores on some metrics such as ROUGE-L, but has lower scores on other metrics than PGN and Onmt-brnn. Our model, however, takes advantage of interaction information from utterances and perform the best on all the metrics (it achieves an improvement of 9.8\% and 10.8\% in terms of the CIDEr metric compared with PGN and Onmt-transformer model respectively.), even though most of the components of our model are similar to the transformer. We also tried to apply coverage mechanism into all approaches, but we found it hurt the performance tremendously.

We now analyze the influence of the beam search size $K$ in the test stage. We contrast the Onmt-brnn model with our model with the beam size in the range of $\{1, 2, 4, 5, 7, 9, 10 \}$. The results are depicted in Figure 3. We can see our model need a bigger beam size to reach the optimal results than Onmt-brnn, and is more stable. We suppose our model need a larger space to search since is use transformer to decode instead of RNN. 

\subsubsection{Qualitative Analysis}
Given two utterances from different two speakers:

Utterance A: \textit{how many are there}

Utterance B: \textit{several, it is in a parking garage at least 10}
Figure 4 visualize the attached co-attention in the utterance interaction layer, consisting of two parts: (1) words of utterance A attends to utterance B; (2) words of utterance B attends to utterance A. We can find in the left part, \textit{serveral}, \textit{at}, \textit{least}, and \textit{10} are paid more attention when \textit{many} is the query. It is reasonable because those number-relevant words are exactly the answers of \textit{how many}. On the other hand, when \textit{several} or other words are treated as queries, \textit{many} is paid more attention because \textit{many} is the most informative word in utterance A.

Here we also provide one quality example of our experiments shown in Figure 5. The dialogue consists of 10 turns and most of them are question-answer pairs. Next to the dialogue is the attached image from MSCOCO dataset(notice the image is placed here just for better understanding the case, we do not involve any image in our setting). Several descriptions, including ground-truth and some system outputs, are placed below. The speakers are talking about a man sitting behind a table, where a group of different wines and a plate of grapes are placed. Speaker A keeps asking the details of the given picture while speaker B keeps answering those questions. However, different from question answering, the questions are popping up sequentially and each question may contain some word like \textit{they}(Turn 2, A) to connect previous questions. 
From the results, we can find all the neural generative models perform well and generate a decent description of the dialogue. As we can summarize, the dialogue contains some key information: a man, a table, wine glasses with different wine, a plate of grapes. All of the generated descriptions are missing grapes information. Attn-Seq2seq and PGN generate some redundant or wrong pieces of information such as \textit{woman}. Onmt-brnn does not mention wine glasses but misuses the word \textit{plate}. Attn-Seq2seq, PGN, and Onmt-transformer miss the information of wines in wine glasses. The description generated by our model, however, is more accurate and comprehensive.

\section{Conclusion}
In this work, we propose a new task named \textit{Dial2Desc}, which encode the input dialogue and decode a high-abstractive-level description. Unlike the previous conversation summarization tasks, we focus more on the object or the action which the speakers are talking about, instead of maintaining the natural flow of the given conversations. We link two open source dataset to create a well-aligned dialogue-to-description dataset \textit{Dial2Desc}. Furthermore, we propose a novel neural attentive model, including an enhanced interaction dialogue encoder and transformer-pointer generator. Results on our \textit{Dial2Desc} dataset demonstrated the effectiveness of our proposed method.

\bibliography{dial2desc_aaai}
\bibliographystyle{aaai}
\end{document}